\documentclass{article} 
\usepackage{iclr2026_conference,times}

\usepackage{amsmath,amsfonts,bm}

\def\eqref#1{equation~\ref{#1}}

\def\1{\bm{1}}

\DeclareMathAlphabet{\mathsfit}{\encodingdefault}{\sfdefault}{m}{sl}
\SetMathAlphabet{\mathsfit}{bold}{\encodingdefault}{\sfdefault}{bx}{n}

\usepackage{hyperref}
\usepackage{url}

\usepackage{graphicx}
\usepackage{subcaption}
\usepackage{lscape}
\usepackage{booktabs}
\usepackage{tabularx}
\usepackage{makecell}
\usepackage{array}
\usepackage{threeparttable}
\usepackage{siunitx}
\usepackage{xcolor}
\usepackage{caption}

\usepackage{amssymb}    
\usepackage[ruled,vlined]{algorithm2e}
\usepackage{float}

\usepackage{wrapfig}

\title{Rewarding the Journey, Not Just the Destination: A Composite Path and Answer Self-Scoring Reward Mechanism for Test-Time Reinforcement Learning}

\author{Jingyu Xing \\
Department of Computer Science\\
Sichuan University\\
Chengdu, 610065, China \\
\texttt{xingjingyuok@stu.scu.edu.cn} \\
\And
Chenwei Tang  \\
Department of Computer Science\\
Sichuan University \\
Chengdu, 610065, China  \\
\texttt{tangchenwei@scu.edu.cn} \\
\And
Xinyu Liu \\
Department of Computer Science\\
Sichuan University\\
Chengdu, 610065, China \\
\texttt{xinyuliu0804@stu.scu.edu.cn} \\
\And
Deng Xiong \\
Stevens Institute of Technology \\
One Castle Point on Hudson \\
Hoboken, NJ 07030-5906, USA \\
\texttt{dxiong@stevens.edu} \\
\And
Shudong Huang \& Wei Ju \& Jiancheng Lv \\
Department of Computer Science\\
Sichuan University\\
Chengdu, 610065, China \\
\texttt{\{huangsd, juwei, lvjiancheng\}@scu.edu.cn} \\
\And
Ziyue Qiao \\
Great Bay University \\
Guangzhou-Hong Kong-Macau Greater Bay Area \\
Guangzhou, China \\
\texttt{ziyuejoe@gmail.com} 
}

\iclrfinalcopy 
\begin{document}

\maketitle

\begin{abstract}
Reinforcement Learning (RL) has emerged as a powerful paradigm for advancing Large Language Models (LLMs), achieving remarkable performance in complex reasoning domains such as mathematics and code generation. However, current RL methods face a fundamental scalability bottleneck due to their heavy reliance on human-curated preference data or labeled datasets for reward modeling. To overcome this limitation, we explore RL on unlabeled data where models learn autonomously from continuous experience streams. The core challenge in this setting lies in reliable reward estimation without ground-truth supervision. Existing approaches like Test-Time RL address this through self-consistent consensus, but risk reinforcing incorrect pseudo-labels derived from majority voting. We introduce \textbf{COMPASS} (\textbf{Com}posite \textbf{P}ath and \textbf{A}nswer \textbf{S}elf-\textbf{S}coring), a novel test-time reward mechanism that operates without external supervision. \textbf{COMPASS} integrates two complementary components: the Dual-Calibration Answer Reward (DCAR), which stabilizes training by establishing trustworthy pseudo-labels through confidence and credibility calibration, and the Decisive Path Reward (DPR), which directly optimizes the reasoning process quality beyond mere outcome supervision. By jointly reinforcing trustworthy consensus answers and highly decisive reasoning chains, the \textbf{COMPASS} systematically enhances the model's analytical capabilities. Extensive experiments show that \textbf{COMPASS} achieves significant and consistent performance gains across diverse reasoning tasks and model architectures, advancing a more scalable direction for LLMs to learn from continuous experience.
\end{abstract}

\section{Introduction}
Reinforcement Learning (RL) \cite{rl,rlhf3} has emerged as a powerful paradigm for advancing the capabilities of pre-trained Large Language Models (LLMs), driving significant progress in complex reasoning domains, e.g., mathematics \cite{rlmath,rlmath2,rlmath3} and code generation \cite{deepseekr1,gpt4o,qwq,rlcode,rlcode2}. However, current RL approaches predominantly rely on explicit external supervision through ground-truth labels \cite{rlvr,rlhf,grpo} or human preference data to construct reward functions. This dependency creates a fundamental scalability bottleneck: as tasks grow in complexity and volume, large-scale annotation becomes increasingly impractical, hindering the continuous evolution of state-of-the-art models.

This limitation naturally motivates an alternative paradigm where LLMs autonomously improve through RL on unlabeled data, learning directly from continuous experience streams \cite{selfreward,selfreward2,selfreward3}. However, the core challenge in this paradigm lies in reward estimation during inference without access to ground-truth information. Test-Time Reinforcement Learning (TTRL) \cite{ttrl} formalizes this setting by enabling parameter updates using unlabeled test data—a promising direction that has recently gained significant attraction. TTRL addresses the reward challenge by sampling multiple responses per problem and constructing pseudo-labels through self-consistency consensus via majority voting. The approach relies on the model's intrinsic confidence as a proxy metric when external rewards are unavailable: answers consistently reproduced across multiple trials are considered higher-confidence and thus more likely to be correct. This mechanism mirrors human problem-solving, where conclusions verified through diverse methods strengthen our confidence in their validity.

While using confidence as a correctness proxy aligns with cognitive principles, a critical question remains: \textbf{\textit{how is confidence actually manifested in the reasoning process of LLMs?}} TTRL\cite{ttrl} adopts self-consistency via majority voting as its confidence proxy. However, this choice has clear limitations. When initial pseudo-labels derived from voting are biased or incorrect, reflecting flaws in the pre-trained model's prior knowledge, the model risks reinforcing erroneous consensus. This raises an important motivation for our work: \textit{\textbf{can we exploit other forms of prior knowledge within LLMs to mitigate these limitations and obtain more reliable self-reward signals?}} Since LLMs fundamentally operate as next-token predictors, their most direct internal state evidence resides in the probability distribution over candidate tokens. Different measures of this distribution naturally yield alternative notions of model confidence. For instance, the entropy of the distribution reflects uncertainty—lower entropy indicates higher confidence. Similarly, the probability assigned to the $\text{top-1}$ token represents the model's certainty in its preferred choice. Moreover, the margin between the $\text{top-1}$ and $\text{top-2}$ token probabilities captures decisiveness; a larger margin implies the model is less hesitant, and thus more confident.

Our analysis indicates that the token-level probability distribution encodes rich internal signals that extend beyond simple majority voting, providing a principled basis for guiding the model’s optimization toward more reliable reasoning and answers. To systematically leverage these signals, we introduce \textbf{COMPASS} (\textbf{Com}posite \textbf{P}ath and \textbf{A}nswer \textbf{S}elf-\textbf{S}coring), a novel reward mechanism that combines answer-level calibration with path-level evaluation. \textbf{COMPASS} consists of two components: the Dual-Calibration Answer Reward (DCAR) and Decisive Path Reward (DPR). First, DCAR refines majority voting by incorporating confidence measures and assessing the credibility of pseudo-labels, effectively enhancing learning stability and efficiency. Furthermore, DPR moves beyond the final answer to scrutinize each step of the generation process. Through an entropy-weighting mechanism, it encourages the model to make more decisive choices (high decisiveness) at critical junctures of high uncertainty (high entropy), providing a direct and dense supervisory signal for optimizing the reasoning path. Our contributions can be summarized as follows:

\begin{itemize}
    \item We propose \textbf{COMPASS}, a novel self-scoring reward mechanism for reinforcement learning on unlabeled data that enables LLM self-evolution through intrinsic evaluation of both final answers and intermediate reasoning paths.

     \item We design a composite reward function featuring two innovative components: DCAR, which dual-calibrates consensus for more reliable answer rewards, and DPR, which introduces process-centric evaluation through dense rewards for decisive token generation during uncertain reasoning steps.

     \item Extensive experiments across diverse reasoning benchmarks demonstrate \textbf{COMPASS}'s effectiveness and superiority, marking a significant advancement in learning from continuous experience streams.
\end{itemize}

\section{Related Work}
\subsection{RL for Reasoning}
RL plays a critical role in enhancing the instruction-following and reasoning capabilities of LLMs \cite{deepseekr1,rlhf}. Over time, research in this direction has evolved through three main paradigms, each progressively reducing reliance on external supervision. Reinforcement Learning from Human Feedback (RLHF) \cite{rlhf,rlhf2,saferlhf}, aligns base models with human preferences using learnable reward models trained on annotated preference data. Methods such as Proximal Policy Optimization (PPO) \cite{ppo} are widely employed, but the dependence on large-scale human feedback limits scalability. Reinforcement Learning with Verifiable Rewards (RLVR) \cite{rlvr,grpo}, replaces preference-based supervision with rule-based reward functions grounded in gold-standard answers. This class of methods, which has been particularly effective for math and code generation tasks, reduces the need for nuanced human feedback but still requires labeled datasets to provide ground-truth answers. Reinforcement Learning from Internal Feedback (RLIF) \cite{empo,ttrl,rlif1}, eliminates explicit supervision altogether by leveraging intrinsic signals from LLMs themselves. The underlying idea is that LLMs already possess a substantial amount of knowledge, which—if effectively organized—can support correct reasoning. However, direct inference after supervised training often fails to fully exploit this potential, requiring reinforcement learning to discover strategies that better coordinate and utilize existing knowledge. Conceptually, RLIF aims to formalize this intuition by framing the organization and utilization of pre-existing knowledge within LLMs as a policy optimization problem, wherein intrinsic feedback signals serve as implicit rewards guiding the emergence of more effective reasoning strategies. For example, EMPO \cite{empo} incentivizes reasoning by minimizing entropy in a latent semantic space, while TTRL \cite{ttrl} generates pseudo-labels via self-consistency consensus among multiple sampled responses and uses them to compute rewards. These approaches mark a shift toward self-rewarding reinforcement learning, aiming to enable LLMs to improve directly from unlabeled data. Our work also falls within the scope of RLIF.

\subsection{Confidence-Based Reward}
In fully unsupervised settings \cite{unsupervised1,unsupervised2}, where ground-truth labels are unavailable, directly optimizing for the correctness of Large Language Model (LLM) outputs is infeasible. A natural alternative is to exploit \textit{intrinsic proxy metrics} that correlate strongly with correctness. Among these, \textit{model confidence} \cite{confidence1,confidence2,confidence3} has emerged as a particularly informative indicator, mirroring human cognition, in which confidence serves as an internal estimate of reliability in the absence of external supervision. Current confidence-based reward methods have evolved along three complementary trajectories. One paradigm quantifies confidence through the \textit{log-likelihood} of generated sequences, rewarding responses that the model assigns higher probability; for instance, RLSC \cite{confidence-is-all-you-need} reinforces generation by maximizing sequence log-likelihood. Another paradigm centers on \textit{entropy minimization}, where lower entropy implies greater certainty—an idea instantiated by RENT \cite{min-entropy}, which encourages more decisive and coherent outputs. A third direction evaluates \textit{self-consistency} across multiple samples, as in TTRL \cite{ttrl}, which derives pseudo-labels via consensus among sampled responses and reinforces outputs consistent with the emergent majority. While these confidence-based methods have yielded notable gains in reasoning and stability, they predominantly rely on a single dimension of confidence—whether likelihood, entropy, or consensus. This leaves open an important question: how can multiple complementary confidence signals be integrated to construct more reliable and fine-grained self-reward mechanisms? Our work addresses this gap by proposing a unified reward framework that leverages diverse intrinsic feedback sources to guide more robust self-improvement in LLMs.

\subsection{Test-Time Adaptation}
Test-Time Adaptation (TTA) \cite{tta1,tta2,tta3} refers to updating a model on unlabeled test data during inference, aiming to mitigate performance degradation caused by distribution shifts between training and testing environments. Early studies in computer vision primarily explored unsupervised objectives that promote robustness to domain shifts. More recently, the TTA paradigm has been extended to LLMs and reasoning tasks. For instance, Tent \cite{tent} adapts models by minimizing the entropy of predictions, under the assumption that well-adapted models should produce certain (i.e., low-entropy) outputs on test samples. Similarly, TTRL \cite{ttrl} introduces test-time reinforcement learning, where rewards are derived from self-consistency via majority voting among sampled responses. Despite these advances, most existing TTA approaches remain limited in scope: they either focus on generic uncertainty reduction or rely on outcome-level consensus, offering only indirect supervision over the underlying reasoning process. This limitation highlights the need for more expressive test-time reward mechanisms that can jointly assess both answer reliability and reasoning quality, thereby enabling stronger and more principled adaptation in LLMs.

\begin{figure}[t]
\centering
\includegraphics[width=\linewidth]{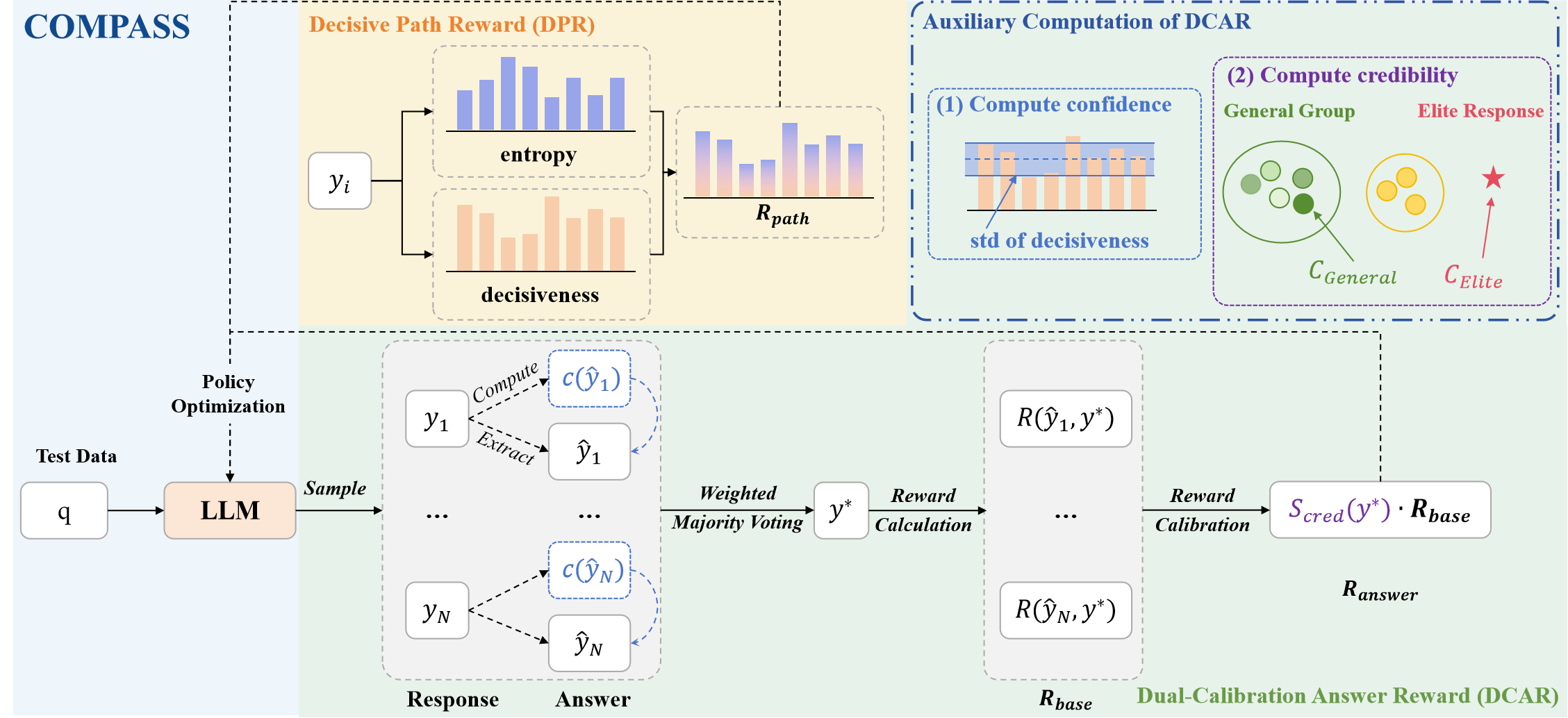}
\caption{The \textbf{Composite Path and Answer Self-Scoring (COMPASS)} reward mechanism. Given a prompt $q$, the policy LLM samples multiple candidate responses $\{y_1, y_2, \cdots, y_N\}$. The \textbf{Dual-Calibration Answer Reward (DCAR)} firstly constructs a consensus pseudo-label $y^*$ via \textit{confidence-calibrated self-consistency}, where each response’s contribution is weighted by its decisiveness-derived confidence. DCAR further evaluates the credibility of this consensus by comparing the confidence of the general supporting group against the most confident elite response, yielding a \textit{reliability-calibrated answer reward} $R_{answer}$. Complementarily, the \textbf{Decisive Path Reward (DPR)} assesses the reasoning process quality by computing step-wise entropy and decisiveness measures to generate the path reward $R_{path}$. Together, DCAR and DPR form a unified intrinsic reward framework that reinforces both trustworthy answers and high-quality reasoning processes.}
\label{fig:NET}
\end{figure}

\section{Composite Path and Answer Self-Scoring}
This section introduces the proposed \textbf{COMPASS} framework for reinforcement learning on unlabeled data. COMPASS is motivated by the observation that the token-level probability distribution of LLMs encodes multiple internal signals, e.g., \textit{uncertainty, confidence, and decisiveness}, which can serve as intrinsic feedback for guiding optimization. COMPASS systematically integrates these signals to provide more reliable and fine-grained rewards. Specifically, given a state represented by the prompt $q$, the model acts by producing an output $y$ sampled from a policy ${\pi }_{\theta }\left( {y \mid  q}\right)$ parameterized by $\theta$. To construct reward signals without ground-truth labels, we generate multiple candidate responses and extract the corresponding answers $\left\{  {\hat{y}_{1},\hat{y}_{2},\ldots ,\hat{y}_{N}}\right\}$ from the model through repeated sampling. A consensus output ${y}^{ * }$ is derived through \textit{\textbf{confidence-calibrated self-consistency}}, serving as a proxy for the optimal action. The environment then provides a reward $R\left( {\hat{y}_i,{y}^{ * }}\right)$ based on the alignment between the sampled action $\hat{y}_i$ and the consensus action ${y}^{ * }$. To further stabilize RL training, we propose a \textit{\textbf{credibility}} metric to assess the quality of generated pseudo-labels. In addition to the aforementioned outcome-based Dual-Calibration Answer Reward (DCAR), we also introduce the process-based Decisive Path Reward (DPR) to evaluate the reasoning quality of each candidate response. As shown in Figure \ref{fig:NET}, the proposed \textbf{COMPASS} consisting of DCAR and DPR achieves reinforcement learning on data without explicit labels. 

\begin{figure}[t]
    \captionsetup[subfigure]{font=scriptsize} 
    \centering
    \begin{subfigure}[b]{0.32\linewidth}
        \centering
        \includegraphics[width=\linewidth]{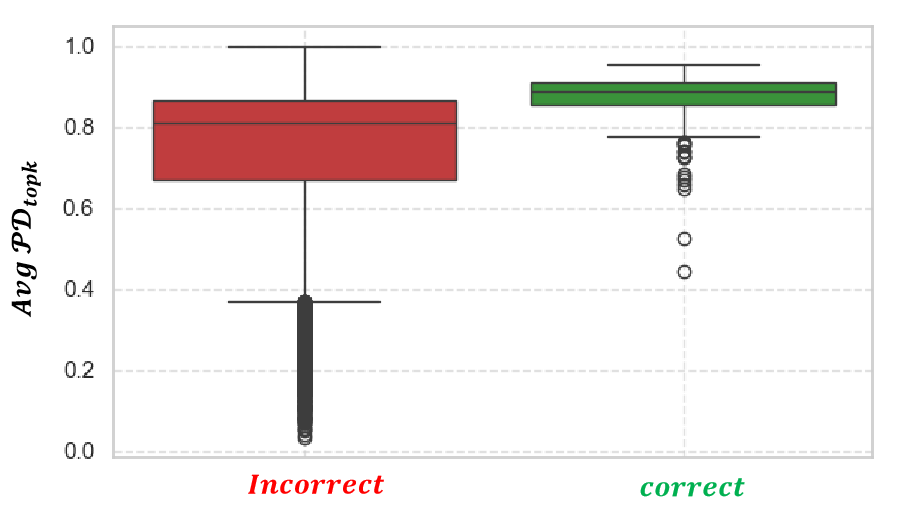}
        \caption{Distribution of average $\mathcal{PD}_{topk}$.}
        \label{fig:avg_topk_box}
    \end{subfigure}
    \hfill
    \begin{subfigure}[b]{0.32\linewidth}
        \centering
        \includegraphics[width=\linewidth]{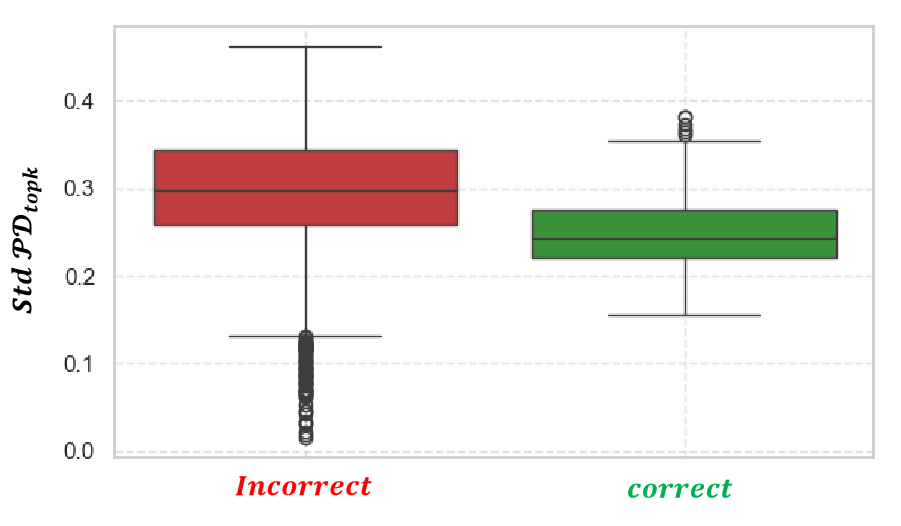}
        \caption{Distribution of $\mathcal{PD}_{topk}$ stability.}
        \label{fig:std_topk_box}
    \end{subfigure}
    \hfill
    \begin{subfigure}[b]{0.32\linewidth}
        \centering
        \includegraphics[width=\linewidth]{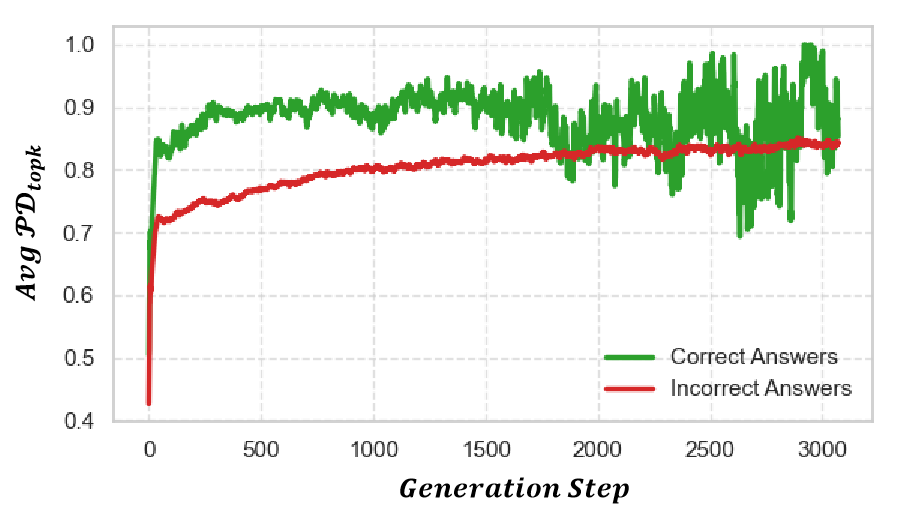}
        \caption{Average $\mathcal{PD}_{topk}$ curves.}
        \label{fig:avg_topk_curve}
    \end{subfigure}
    
    \vspace{-0.5em}
    \caption{Analysis of $\mathcal{PD}_{topk}$ indicators validating confidence-calibrated self-consistency. (a) Distribution of average $\mathcal{PD}_{topk}$ values, showing systematic differences between correct and incorrect responses. (b) Distribution of $\mathcal{PD}_{topk}$ stability throughout generation trajectories, demonstrating more consistent decisiveness in correct answers. (c) Average $\mathcal{PD}_{topk}$ curves across generation steps, revealing that correct responses maintain higher and more stable decisiveness throughout reasoning. These empirical findings confirm that the probability difference between top-1 and top-2 tokens strongly correlates with answer correctness.}
    \label{fig:topk_diff_three}
\end{figure}

\subsection{Dual-Calibration Answer Reward (DCAR)}
A central challenge in RL on unlabeled data is ensuring the reliability of pseudo-labels for stable optimization. To address this, we design the DCAR, which refines consensus answers by integrating two complementary signals: \textit{confidence} and \textit{credibility}. Our approach begins with a key hypothesis: \textit{more confident responses should contribute more significantly to the final decision}. As shown in Figure \ref{fig:topk_diff_three}, this hypothesis is supported by our correlation analysis, which reveals that $\mathcal{PD}_{topk}$, i.e., the probability difference between top-1 and top-2 tokens across the generation trajectory, strongly correlates with final answer correctness, capturing the model's predictive stability and decisiveness. Building on this insight, we implement \textit{confidence-calibrated self-consistency}, where more decisive responses receive higher weighting in pseudo-label formation. Therefore, we first define the \textit{confidence} of a trajectory $\hat{y}_i$ as:

\begin{equation}
\mathcal{PD}_{topk}(x_t)=p(x_t^1 | x_{<t}) - p(x_t^2 | x_{<t}),
\end{equation}
\begin{equation}
\text{c}(\hat{y}_i) = \exp \left( - \text{std}_t \ \mathcal{PD}_{topk}(x_t) \right),
\end{equation}
where $x_t^1$ and $x_t^2$ represent the top-1 and top-2 tokens at timestep $t$. And we then define \textit{confidence} as the negative exponential of the standard deviation ($std$), which ensures lower $std$ yields higher confidence and guarantees positive weights required for self-consistency. Finally, for a given answer $y$, we design the \textit{confidence-calibrated self-consistency} score $\mathcal{S}_{ccsc}$ as the confidence-weighted sum:
\begin{equation}
    \mathcal{S}_{ccsc}(y) = \frac{\sum_{i: \hat{y}_i = y} \text{c}(\hat{y}_i)}{\sum_{i=1}^{N} \text{c}(\hat{y}_i)}.
\end{equation}
The final pseudo-label $y^*$ is determined as the answer with the highest calibrated score:
\begin{equation}
y^* = \arg\max_{y} \mathcal{S}_{ccsc}(y).
\end{equation}

To further enhance the reliability of our confidence-calibrated pseudo-label $y^*$, we introduce a \textit{credibility} score $\mathcal{S}_{cred}$ that assesses consensus quality relative to the most confident response, effectively implementing soft curriculum learning. This approach is grounded in a key hypothesis: \textit{a consensus derived from high-confidence responses is more reliable than one based on diverse low-confidence outputs}. Building on this principle, we define the credibility metric using two fundamental concepts: \textit{General Group} $\mathcal{C}_{\text{General}}$ and \textit{Elite Response} $\mathcal{C}_{\text{Elite}}$. Specifically, the \textit{General Group} contains all responses that agree with the pseudo-label $y^*$. The group's confidence $\mathcal{C}_{\text{General}}$ is defined as the maximum confidence within this subset, representing the strongest supporting evidence for the consensus, and the $\mathcal{C}_{\text{Elite}}$ is the response among all $N$ candidates with the highest confidence:
\begin{equation}
    \mathcal{C}_{\text{General}} = \max_{i: \hat{y}_i = y^*} c(\hat{y}_i), \quad \mathcal{C}_{\text{Elite}} = \max_{i=1, \ldots, N} c(\hat{y}_i).
\end{equation}
The \textit{credibility} of the pseudo-label $y^*$ is the ratio of these two confidence scores:
\begin{equation}
\mathcal{S}_{cred}(y^*) = \frac{\mathcal{C}_{\text{General}}}{\mathcal{C}_{\text{Elite}}}.
\end{equation}
This ratio quantifies the consensus strength relative to the most confident individual response. A value of 1 indicates perfect alignment between the consensus and the most confident opinion, signifying high reliability, whereas values below 1 reveal the presence of highly confident dissenters, thereby reducing trust in the consensus outcome. The final outcome reward $R_{\text{answer}}$ in our DCAR module integrates the base reward $R_{\text{base}}(\hat{y}_i)$ with the credibility score $\mathcal{S}_{cred}$ through multiplicative modulation:
\begin{equation}
R_{\text{answer}}(\hat{y}_i) = \mathcal{S}_{cred}(y^*) \cdot R_{\text{base}}(\hat{y}_i),
\end{equation}
where $R_{\text{base}}(\hat{y}_i)$ is a binary indicator that equals 1 if $\hat{y}_i$ matches the pseudo-label $y^*$ and 0 otherwise. This formulation transforms the sparse binary reward into a continuous signal within $[0, 1]$, implementing a soft curriculum learning mechanism that directs the model's focus toward high-credibility pseudo-labels, thereby promoting more stable and reliable optimization.
\begin{algorithm}[t]
    \caption{The Complete Composite Path and Answer Self-Scoring Reward.}
    \label{alg:compass}
    \footnotesize
    \KwIn{Prompt $q$, policy $\pi_{\theta}$, number of samples $N$}
    \KwOut{Final rewards $\{R(y_i)\}_{i=1}^N$}
    \textbf{Initialize:} trajectories $Y$, answers $\hat{Y}$, confidences $C$, decisiveness sequences $D$, entropy sequences $H$ \\
    
    \tcc{--------------DCAR: Dual-Calibration Answer Reward--------------}
    \For{$i \leftarrow 1$ \KwTo $N$}{
        Sample $y_i \sim \pi_{\theta}(\cdot | q)$, extract answer $\hat{y}_i$ \\
        Compute confidence $\text{c}(\hat{y}_i) = \exp ( - \text{std}_t \ \cdot \mathcal{PD}_{topk}(x_t) )$ \\
        Append to $Y, \hat{Y}, C$\;
    }
    
    Find unique answers $A = \text{unique}(\hat{Y})$, initialize $S[a] \leftarrow 0$ \\
    \ForEach{$a \in A$}{
        $S[a] \leftarrow \sum_{\hat{y}_i = a} \text{c}(\hat{y}_i)$ 
    }
    $y^* = \arg\max_{a \in A} S[a]$ \\
    $\mathcal{C}_{\text{General}} \leftarrow \max(\{\text{c}(\hat{y}_i) \mid \hat{y}_i = y^*\})$ \\ 
    $\mathcal{C}_{\text{Elite}} \leftarrow \max(\{\text{c}(\hat{y}_i)\})$ \\
    $\mathcal{S}_{cred}(y^*) \leftarrow \mathcal{C}_{\text{General}} / \mathcal{C}_{\text{Elite}}$ \\
    
    \For{$i \leftarrow 1$ \KwTo $N$}{
        $R_{\text{base}} \leftarrow \mathbb{I}[\hat{y}_i = y^*]$ \\
        $R_{\text{answer}}(y_i) \leftarrow \mathcal{S}_{cred}(y^*) \cdot R_{\text{base}}$ 
    }

    \tcc{-------------------DPR: Decisive Path Reward-------------------}
    \For{$i \leftarrow 1$ \KwTo $N$}{
        \For{$t \leftarrow 1$ \KwTo $T$}{
            $d_t \leftarrow \mathcal{PD}_{topk}(x_t)$ \\
            $h_t \leftarrow -\sum_j p(x_t^j|x_{<t}) \log p(x_t^j|x_{<t})$ \\
            Append $d_t$ to $D[i]$, $h_t$ to $H[i]$\ \\
        }
        $w_t \leftarrow \frac{e^{h_t}}{\sum_{j=1}^T e^{h_j}}$  \\
        $R_{\text{path}}(y_i) \leftarrow \sum_{t=1}^T w_t \cdot d_t$ 
    }
    
    \tcc{----------------------COMPASS Integration----------------------}
    \For{$i \leftarrow 1$ \KwTo $N$}{
        $R(y_i) \leftarrow R_{\text{answer}}(y_i) + R_{\text{path}}(y_i)$ 
    }
    \KwRet $\{R(y_i)\}_{i=1}^N$\;
\end{algorithm}

\subsection{Decisive Path Reward (DPR)}
While DCAR provides robust outcome-level supervision, effective reasoning requires complementary process-level guidance. To address this, we introduce the DPR to encourage decisive actions at critical reasoning junctures to ensure the integrity of the entire reasoning chain. DPR operates by providing dense per-token feedback that directly optimizes reasoning trajectories through entropy-weighted decisiveness scoring. Specifically, we evaluate two complementary metrics at each generation step $t$: \textit{decisiveness} $d_t$, quantifying the model's confidence in its token selection, and \textit{uncertainty} $h_t$, measuring the entropy of the predictive distribution. The decisiveness at each position is defined as:
\begin{equation}
d_t=\mathcal{PD}_{topk}(x_t),
\end{equation}
with higher values $\text{d}(x_t)$ indicating more confident and unambiguous decisions. Our central hypothesis posits that \textit{decisiveness carries greater importance during high-uncertainty moments}, where confident actions provide more informative signals when multiple alternatives appear viable. Based on this premise, we define the process reward $R_{\text{path}}(y_i)$ by dynamically weighting each step's decisiveness by its corresponding uncertainty:
\begin{equation}
    w_t = \frac{e^{h_t}}{\sum_{j=1}^{T} e^{h_j}}, \quad R_{\text{path}}(y_i) = \sum_{t=1}^{T} w_t \cdot d_t.
\end{equation}
This formulation provides dense, per-token feedback that incentivizes decisive actions at critical high-uncertainty junctures, fostering more robust reasoning paths. The complete COMPASS reward combines both components (see Algorithm \ref{alg:compass}):
\begin{equation}
R(y_i) = R_{\text{answer}}(y_i) +  R_{\text{path}}(y_i).
\end{equation}

\section{Experiments}
\subsection{Experimental Setup}
\textbf{Benchmarks.} To comprehensively evaluate COMPASS's generalizability, we employ diverse backbone models spanning different scales and specializations: LLaMA-3.2-1B-Instruct \cite{llama3.2} as an instruct-tuned model, Qwen2.5-Math-1.5B \cite{qwen-math-1.5} as a mathematically specialized base model, and Qwen2.5-7B \cite{qwen-math-1.5} as a general-purpose base model. We conduct evaluation on GPQA-Diamond \cite{gpqa}, a challenging subset of the Graduate-Level Google-Proof Question Answering benchmark, alongside three mathematical reasoning benchmarks: AIME 2024 \cite{aime}, AMC \cite{aime}, and MATH-500 \cite{math}. As TTRL \cite{ttrl} pioneered the test-time reinforcement learning paradigm, we primarily compare COMPASS against both backbone models and TTRL baselines to validate its effectiveness in enabling autonomous self-improvement.

\textbf{Metrics.} We evaluate COMPASS independently on each benchmark using the pass@k protocol following DeepSeek-R1 \cite{deepseekr1}, reporting pass@1 scores with non-zero temperature sampling. Specifically, we generate 16 responses per question (4 for 32k context length) using temperature 0.6 and top-$p$ 0.95. The pass@1 score is computed as: $\text{pass@1} = \frac{1}{k} \sum_{i=1}^{k} p_i,$, where $p_i \in {0,1}$ indicates the correctness of the $i$-th response.

\textbf{Implementation Details.} We implement COMPASS using GRPO \cite{grpo} independently on each benchmark. For optimization, we employ AdamW with a cosine learning rate schedule (peak $5\times10^{-7}$). During rollout, we sample 64 responses (temperature 0.6, except 1.0 for Qwen2.5-Math) for pseudo-label estimation via voting, then downsample to 32 responses per prompt for training. This design maintains computational efficiency while achieving strong performance. For models with fewer than 7B parameters, we follow TTRL settings with 10, 30, and 80 episodes for MATH-500, AMC, and AIME 2024 respectively, scaled by dataset size. For Qwen2.5-7B, we reduce epochs to 2, 8, and 20 (approximately 20\% of TTRL's budget) to evaluate efficiency in computationally constrained regimes.
\begin{table*}[t]
    \centering
    \caption{Performance comparison on test-time reinforcement learning on various benchmarks. * indicates our reproduction of TTRL; $\dagger$ denotes an evaluation with a reduced number of training epochs compared to the original TTRL paper, a condition applied to both methods for a fair comparison.}
    \label{tab:main_results}
    \small
    \setlength{\tabcolsep}{5pt} 
    \renewcommand{\arraystretch}{1.0} 
    \begin{threeparttable}
    \begin{tabular*}{\textwidth}{@{\extracolsep{\fill}}lcccc}
    \toprule
    \textbf{Name} & \textbf{AIME} & \textbf{AMC} & \textbf{MATH} & \textbf{GPQA}\\
    \midrule
    \multicolumn{5}{c}{\textbf{Instruct Models}} \\
    \midrule
    \textit{LLaMA3.2-1B-Instruct} & 1.5  & 9.8 & 24.7 & 23.8 \\
    \textit{TTRL} & 6.7  & 19.2 & 27.8 & 24.0 \\
    \textbf{\textit{COMPASS}}  & 3.5  & 20.1 & 28.7 & 25.8 \\
    $\Delta$          & \textcolor{green!60!black}{-3.1} & \textcolor{red}{+0.9} & \textcolor{red}{+0.9} & \textcolor{red}{+1.8}\\
    \midrule
    \multicolumn{5}{c}{\textbf{Math Base Models}} \\
    \midrule
    \textit{Qwen2.5-Math-1.5B} & 7.7  & 28.6 & 32.7 & 24.9 \\
    \textit{TTRL}& 15.8 & 47.4\tnote{*} & 72.4\tnote{*} & 26.1 \\
    \textbf{\textit{COMPASS}}& 18.3 & 48.6 & 73.1 & 29.3 \\
    $\Delta$          & \textcolor{red}{+2.5} & \textcolor{red}{+1.2} & \textcolor{red}{+0.7} & \textcolor{red}{+3.2} \\
    \midrule
    \multicolumn{5}{c}{\textbf{Vanilla Base Models}} \\
    \midrule
    \textit{Qwen2.5-7B}\tnote{$\dagger$} & 7.5  & 34.6 & 60.9 & 30.5 \\
    \textit{TTRL} & 20.0 & 50.2 & 76.6 & 31.1 \\
    \textbf{\textit{COMPASS}}& 23.5 & 53.2 & 76.9 & 31.7 \\
    $\Delta$          & \textcolor{red}{+3.5} & \textcolor{red}{+3.0} & \textcolor{red}{+0.3} & \textcolor{red}{+0.6} \\
    \bottomrule
    \end{tabular*}
    \end{threeparttable}
\end{table*}

\begin{figure}[t]
    \captionsetup[subfigure]{font=scriptsize} 
    \centering
    \begin{subfigure}[b]{0.32\linewidth}
        \centering
        \includegraphics[width=\linewidth]{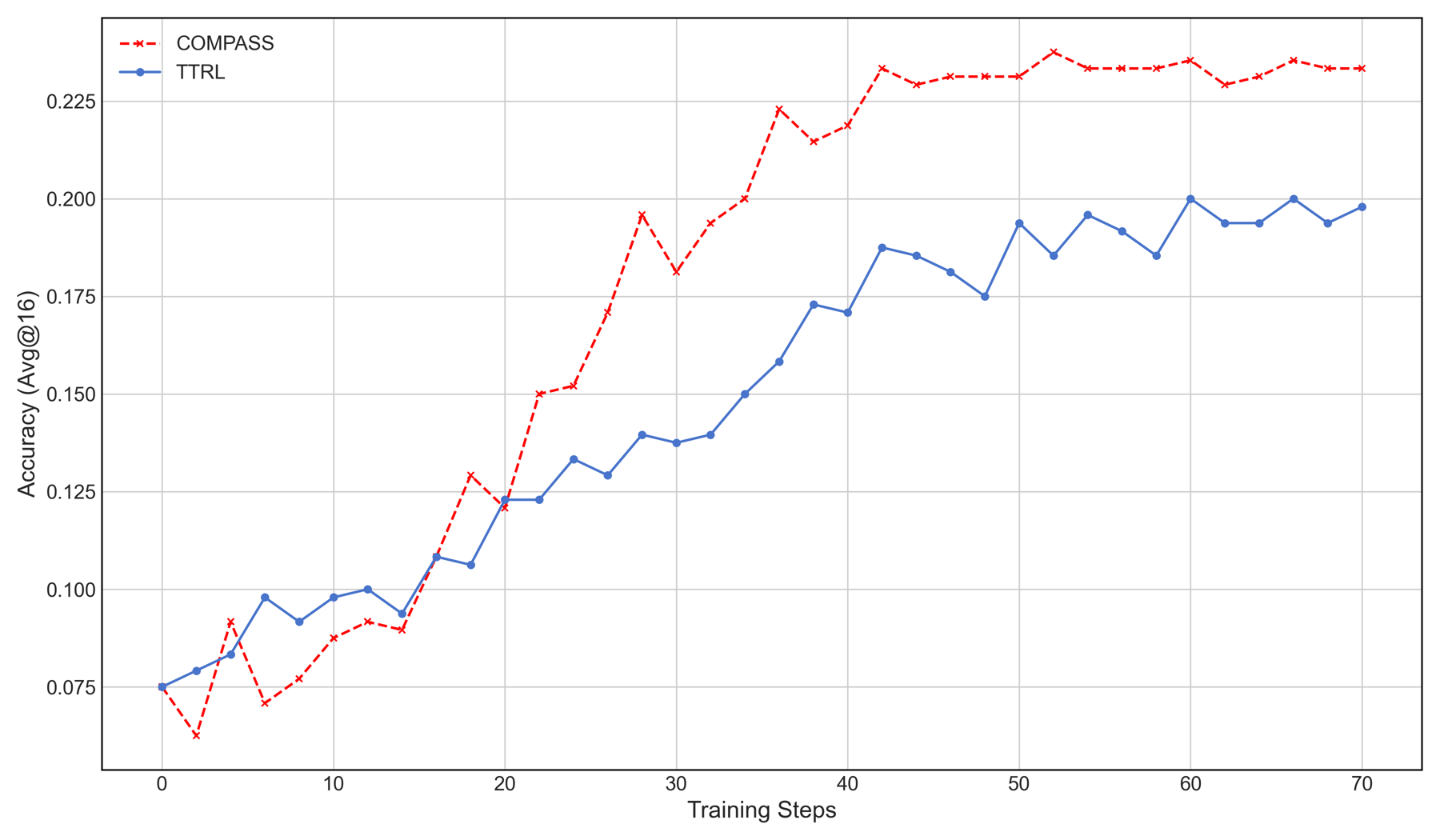}
        \caption{AIME}
        \label{fig:aime}
    \end{subfigure}
    \hfill
    \begin{subfigure}[b]{0.32\linewidth}
        \centering
        \includegraphics[width=\linewidth]{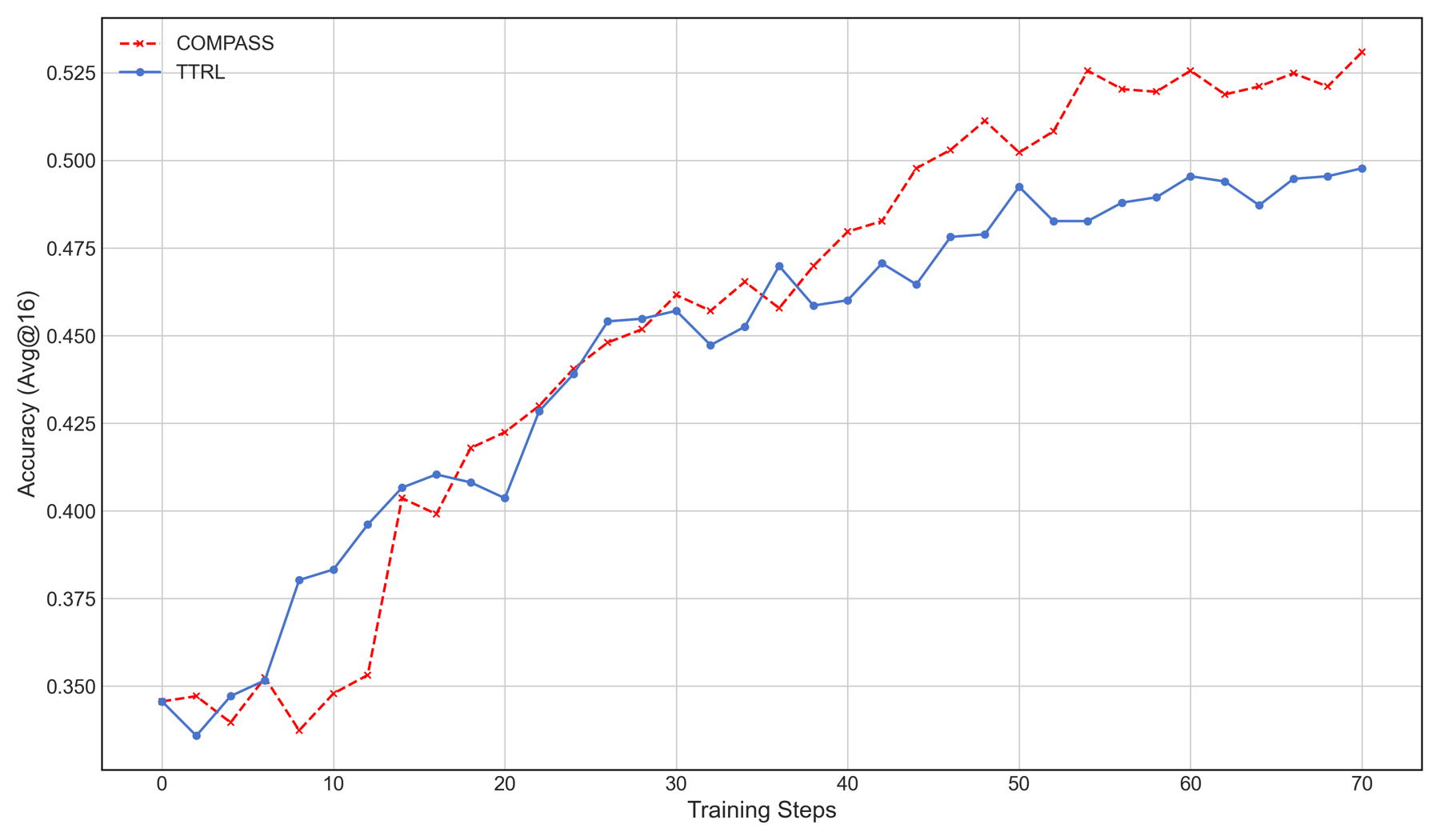}
        \caption{AMC}
        \label{fig:amc}
    \end{subfigure}
    \hfill
    \begin{subfigure}[b]{0.32\linewidth}
        \centering
        \includegraphics[width=\linewidth]{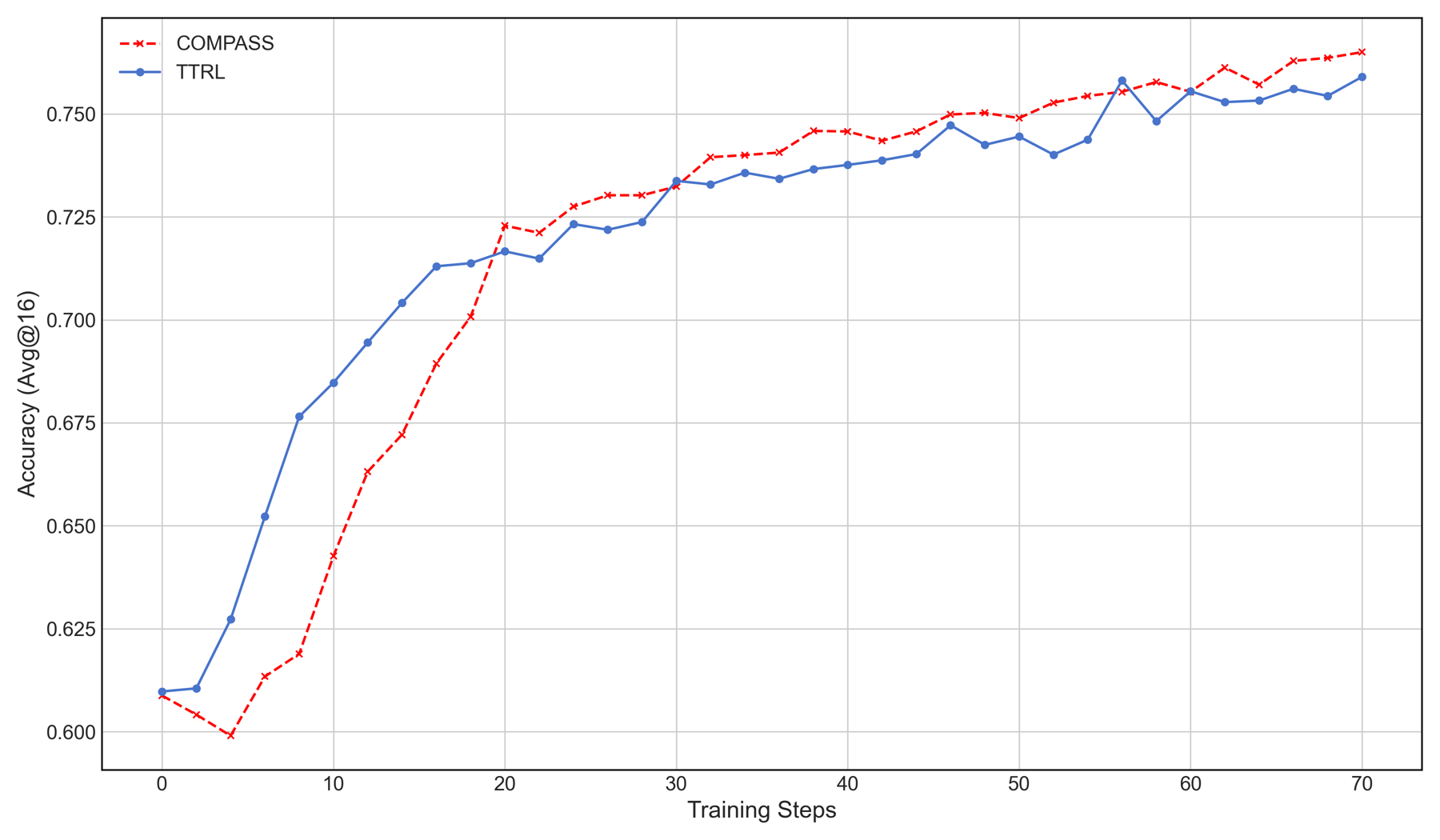}
        \caption{Math}
        \label{fig:math}
    \end{subfigure}
    
    \vspace{-0.5em}
    \caption{
        Performance comparison on AIME/AMC/MATH using Qwen2.5-7B.
    }
    \label{fig:performance-7b}
\end{figure}

\subsection{Main Results}
\textbf{\textit{COMPASS performs well on most tasks and models.}} Table \ref{tab:main_results} presents the main results. We apply COMPASS to 3 models spanning 3 model families, 2 model types, and 3 model sizes, consistently demonstrating obvious improvements across 4 highly challenging benchmarks. On the AIME 2024 and GPQA benchmarks, COMPASS achieves significant improvements of 15.8\% and 12.3\%, respectively, over TTRL when using Qwen2.5-Math-1.5B. For experiments with the larger Qwen2.5-7B base model, both TTRL and COMPASS were trained for approximately 20\% of the epochs specified in the original TTRL paper due to computational constraints. Despite this reduced training schedule, our method demonstrates a clear and consistent performance advantage over TTRL, as evidenced by both the final evaluation metrics in Table \ref{tab:main_results} and the performance trend curves illustrated in Figure \ref{fig:performance-7b}. However, we note an exception with the LLaMA3.2-1B-Instruct model on the AIME 2024 dataset. We attribute this performance drop to the model's insufficient foundational knowledge. For such a model, the high-entropy states targeted by our process reward (DPR) likely signify fundamental confusion rather than meaningful reasoning junctures. Reinforcing these spurious signals inadvertently degrades performance, highlighting that the efficacy of our method relies on the base model possessing a solid knowledge foundation.

\textbf{\textit{COMPASS naturally scales.}} As shown in Table \ref{tab:main_results}, another noteworthy observation is that as the model size increases (1B →1.5B → 7B), performance consistently improves, highlighting the natural scaling behavior of COMPASS: larger models can produce more accurate rewards during self-improvement, which leads to more effective learning on new data.
\begin{figure}[htbp]
\centering
\begin{subfigure}{0.48\linewidth}
  \centering
  \includegraphics[width=\linewidth]{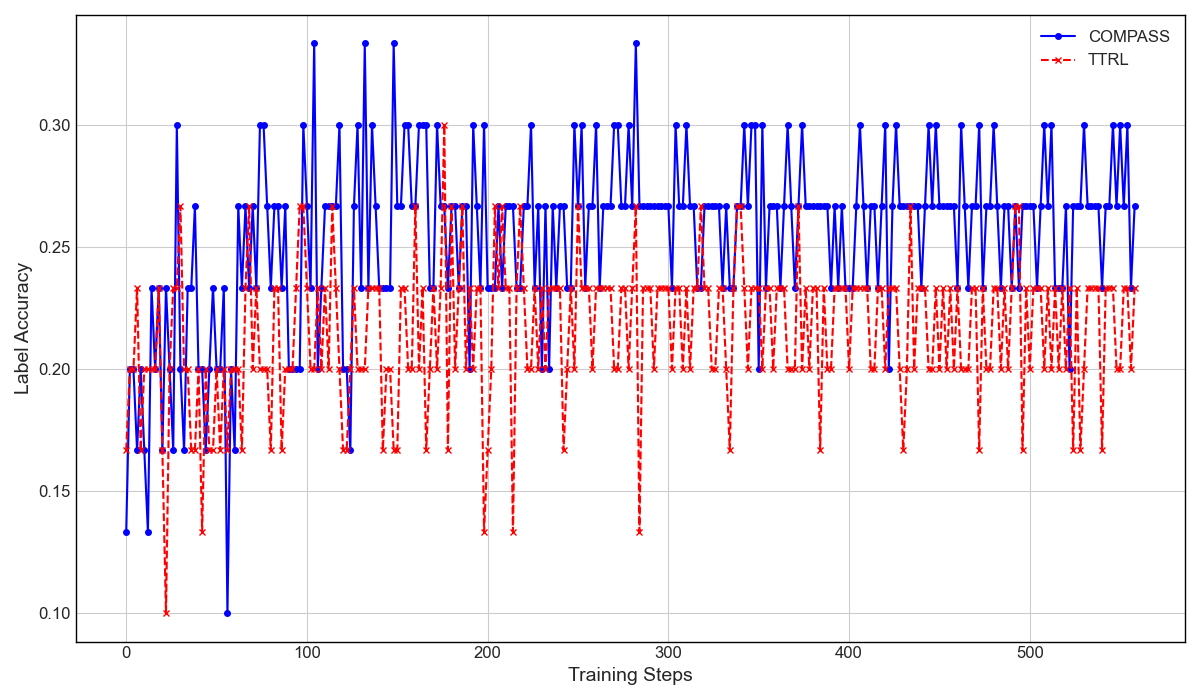}
  \caption{Label Accuracy Curve}
  \label{fig:label_accuracy}
\end{subfigure}
\begin{subfigure}{0.48\linewidth}
  \centering
  \includegraphics[width=\linewidth]{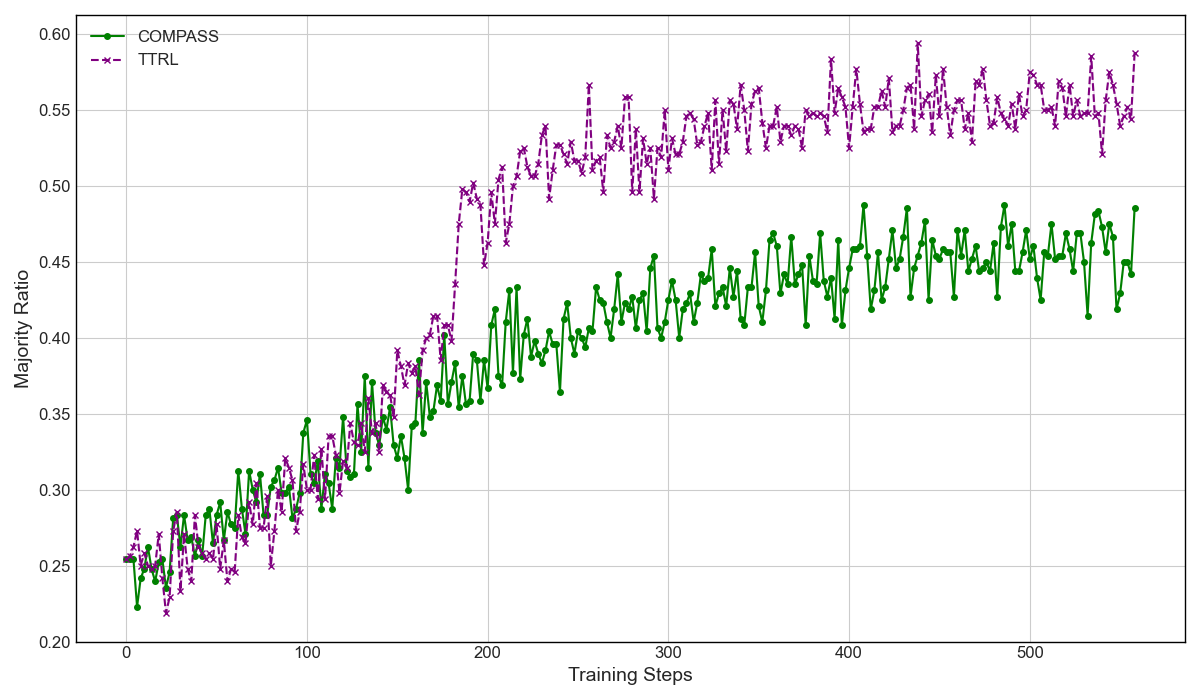}
  \caption{Majority Ratio Curve}
  \label{fig:majority_ratio}
\end{subfigure}
\caption{Training dynamics comparision on AIME using Qwen2.5-Math-1.5B.}
\label{fig:dynamics}
\end{figure}

\textbf{\textit{COMPASS achieves sustainable self-evolution through online and RL.}} To understand the mechanisms of our proposed COMPASS framework, we analyzed its training dynamics against the TTRL baseline as shown in Figure \ref{fig:dynamics}, focusing on\textbf{\textit{ pseudo-label accuracy}} and \textit{\textbf{majority ratio}}. The results highlight COMPASS's superior learning process. Our method achieves significantly higher pseudo-label accuracy, confirming that its advanced reward system—which combines the outcome-based Dual-Calibration Answer Reward (DCAR) and process-based Decisive Path Reward (DPR), and generates more effective training signals. In contrast, the baseline's accuracy stagnates at a much lower level. Simultaneously, COMPASS maintains a consistently lower majority ratio. This demonstrates that it successfully avoids the baseline's tendency to prematurely converge on the most frequent answer, a common pitfall of naive majority voting. Instead of simply reinforcing the consensus, COMPASS values diverse and high-quality reasoning paths. \textit{\textbf{This dual dynamics of increasing label accuracy while reducing reliance on a single popular answer provides strong evidence for COMPASS's effectiveness.}} It cultivates a more robust self-evolution by considering both the intrinsic quality of reasoning paths and the popularity of final answers, leading to a more reliable self-improving model.

\subsection{Ablation Results}
\begin{figure}[!htbp]
\centering
\includegraphics[width=0.85\linewidth]{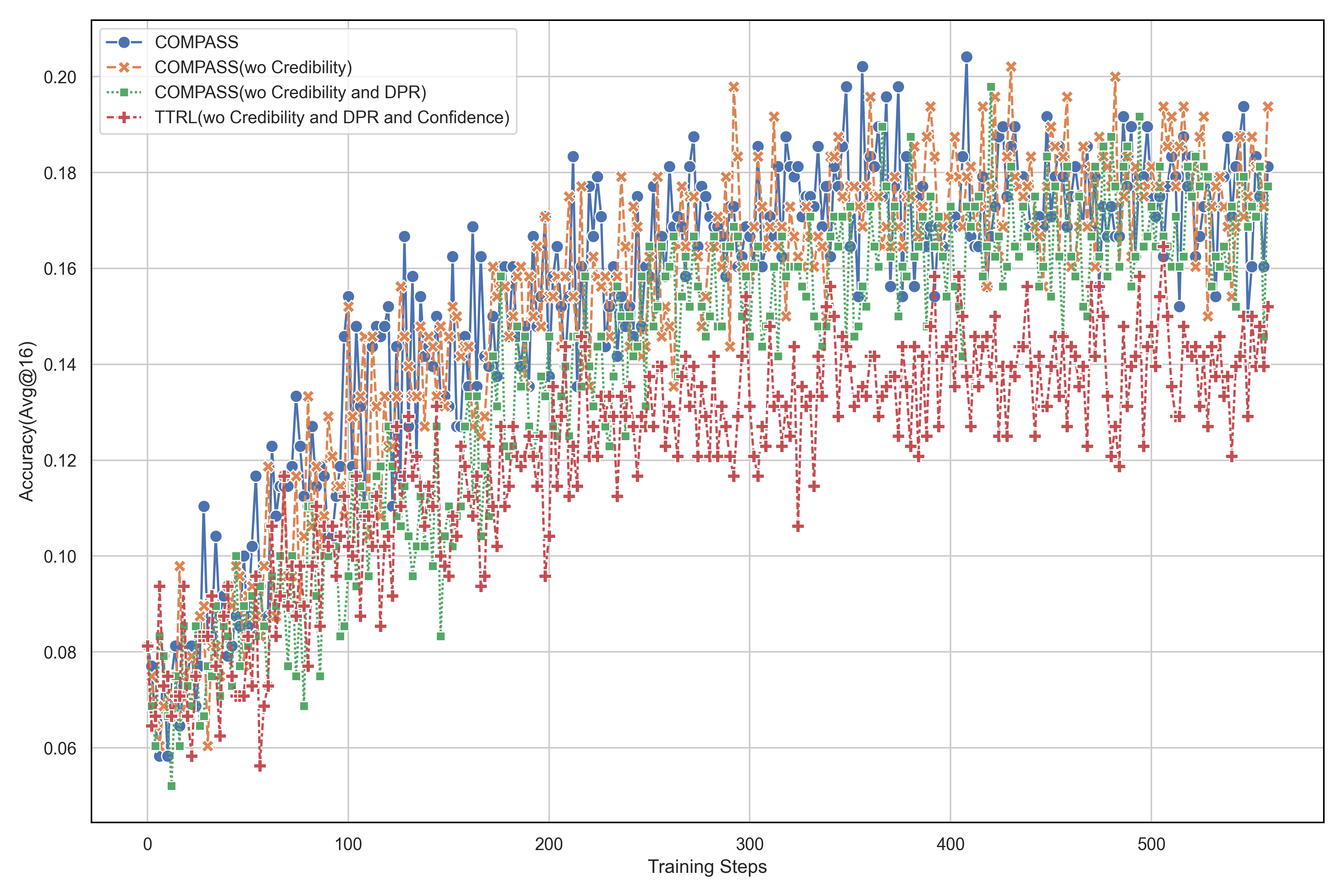}
\caption{Ablation Results of COMPASS on AIME using Qwen2.5-Math-1.5B.}
\label{fig:ablation}
\end{figure}

We performed sequential ablation experiments by progressively removing model components: we first removed the credibility calibration from COMPASS, then further excluded the process reward (DPR), and finally removed the confidence calibration, which reduced the model to the baseline TTRL. The performance curves across training steps (Figure \ref{fig:ablation}) show that each component contributes positively to model performance. Among them, the confidence calibration yields the most substantial improvement, as indicated by the largest vertical gap between the red and green curves.

\subsection{Case Study}
\begin{figure}[!htbp]
\centering
\includegraphics[width=\linewidth]{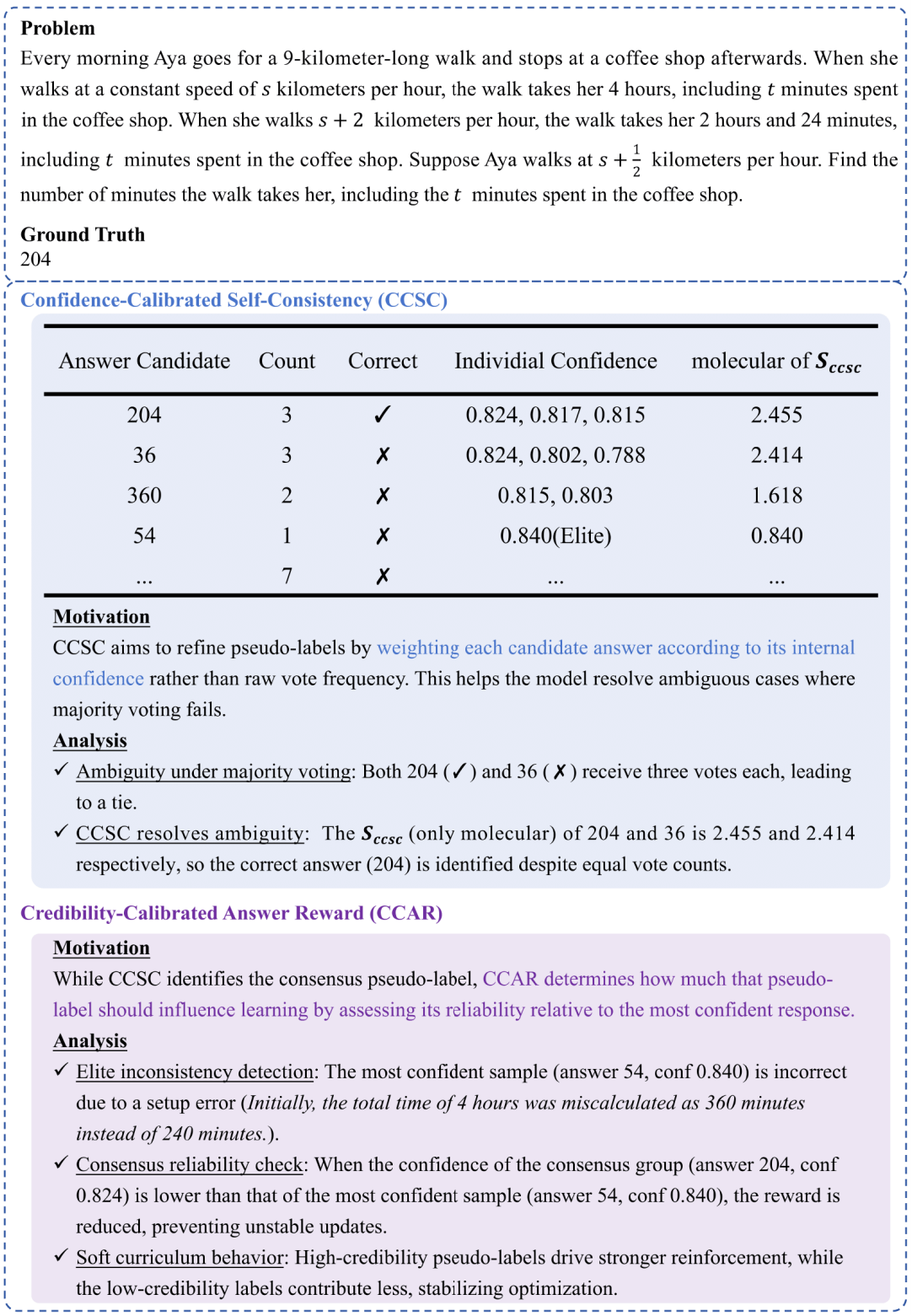}
\caption{Case Study for Dual-Calibration Answer Reward (DCAR)}
\label{fig:case_dcar}
\end{figure}

\begin{figure}[H]
\centering
\includegraphics[width=\linewidth]{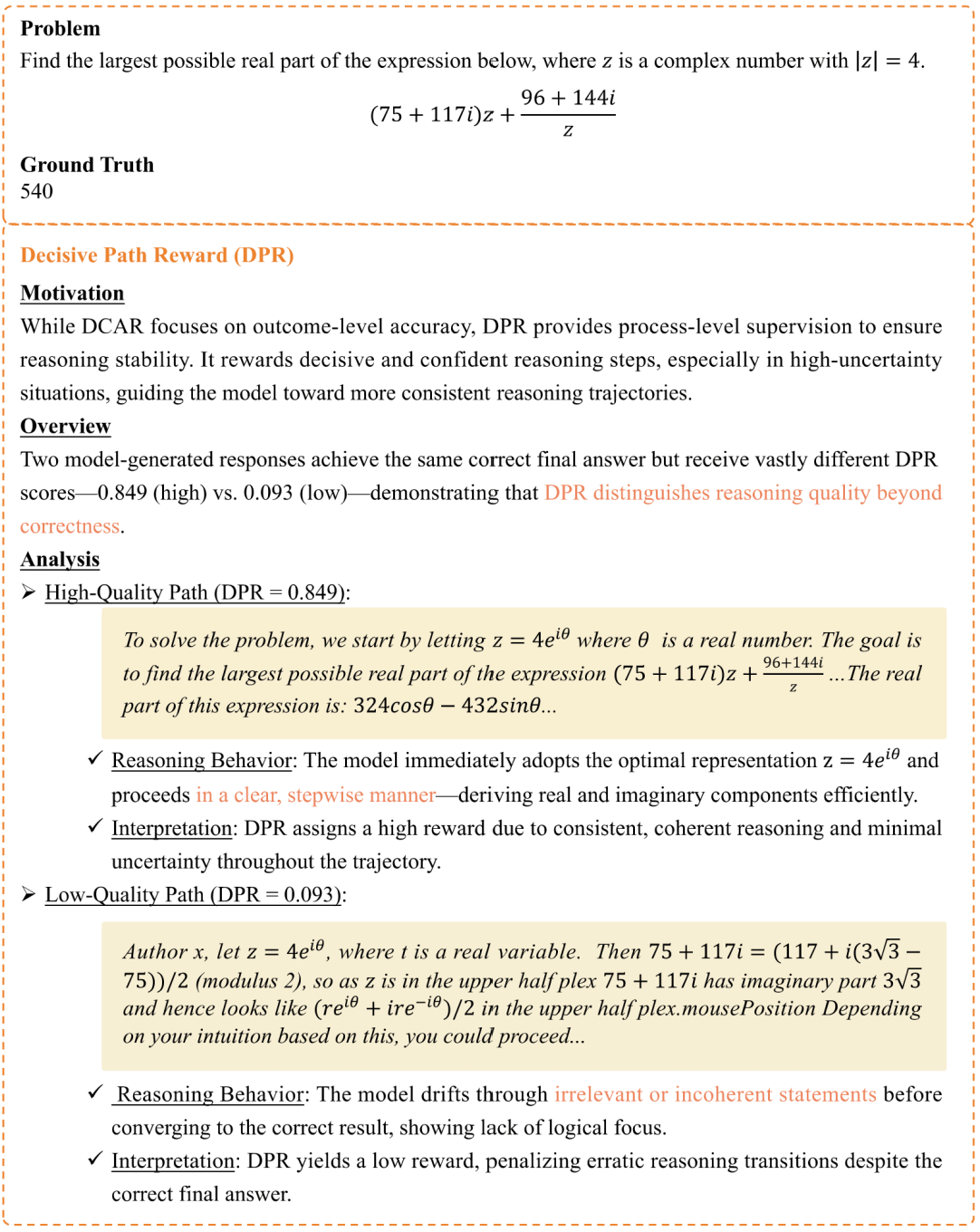}
\caption{Case Study for Decisive Path Reward (DPR).}
\label{fig:case_dpr}
\end{figure}

To validate the effectiveness of COMPASS, we present two case studies—DCAR (Figure~\ref{fig:case_dcar}) and DPR (Figure~\ref{fig:case_dpr}). DCAR establishes trustworthy pseudo-labels through confidence and credibility calibration, while DPR directly evaluates reasoning quality via entropy-weighted decisiveness. Together, these two components demonstrate how COMPASS achieves robust self-evolving reinforcement learning on unlabeled data.
\subsubsection{Dual-Calibration Answer Reward (DCAR)}

\subsubsection{Decisive Path Reward (DPR)}
\section{Conclusion}
In this work, we introduced \textbf{COMPASS}, a novel self-scoring reinforcement learning framework designed to enable Large Language Models to learn from unlabeled data. By integrating two complementary reward components---Dual-Calibration Answer Reward (DCAR) and Decisive Path Reward (DPR)---COMPASS jointly optimizes the reliability of final answers and the decisiveness of reasoning trajectories. Our experiments demonstrate the strong potential of COMPASS, achieving consistent improvements across a variety of tasks and models. We view COMPASS as a further step toward Reinforcement Learning with self-labeled rewards, marking an important direction of learning from continuous streams of self-experience.

\bibliographystyle{iclr2026_conference}

\end{document}